\documentclass[conference]{IEEEtran}
\IEEEoverridecommandlockouts
\usepackage{cite}
\usepackage{amsmath,amssymb,amsfonts}
\usepackage{algorithmic}
\usepackage{graphicx}
\usepackage{textcomp}
\usepackage{xcolor}
\usepackage[utf8]{inputenc}
\usepackage{todonotes}

\def\BibTeX{{\rm B\kern-.05em{\sc i\kern-.025em b}\kern-.08em
    T\kern-.1667em\lower.7ex\hbox{E}\kern-.125emX}}
\begin{document}

\title{Improving Defensive Distillation using Teacher Assistant}

\author{\IEEEauthorblockN{1\textsuperscript{st} 	
Maniratnam Mandal}
\IEEEauthorblockA{\textit{Electrical and Computer Engineering} \\
\textit{Univeristy of Texas at Austin}\\
Austin, USA \\
mmandal@utexas.edu}
\and
\IEEEauthorblockN{1\textsuperscript{st} Suna Guo}
\IEEEauthorblockA{\textit{Computer Science} \\
\textit{Univeristy of Texas at Austin}\\
Austin, USA \\
sguo19@utexas.edu}
}

\maketitle

\begin{abstract}
Adversarial attacks pose a significant threat to the security and safety of deep neural networks being applied to modern applications. More specifically, in computer vision-based tasks, experts can use the knowledge of model architecture to create adversarial samples imperceptible to the human eye. These attacks can lead to security problems in popular applications such as self-driving cars, face recognition, etc. Hence, building networks which are robust to such attacks is highly desirable and essential. Among the various methods present in literature, defensive distillation has shown promise in recent years. Using knowledge distillation, researchers have been able to create models robust against some of those attacks. However, more attacks have been developed exposing weakness in defensive distillation. In this project, we derive inspiration from teacher assistant knowledge distillation and propose that introducing an assistant network can improve the robustness of the distilled model. Through a series of experiments, we evaluate the distilled models for different distillation temperatures in terms of accuracy, sensitivity, and robustness. Our experiments demonstrate that the proposed hypothesis can improve robustness in most cases. Additionally, we show that multi-step distillation can further improve robustness with very little impact on model accuracy.  
\end{abstract}


\section{Introduction and Background}
\begin{figure}
    \centering
    \includegraphics[width=0.9\linewidth]{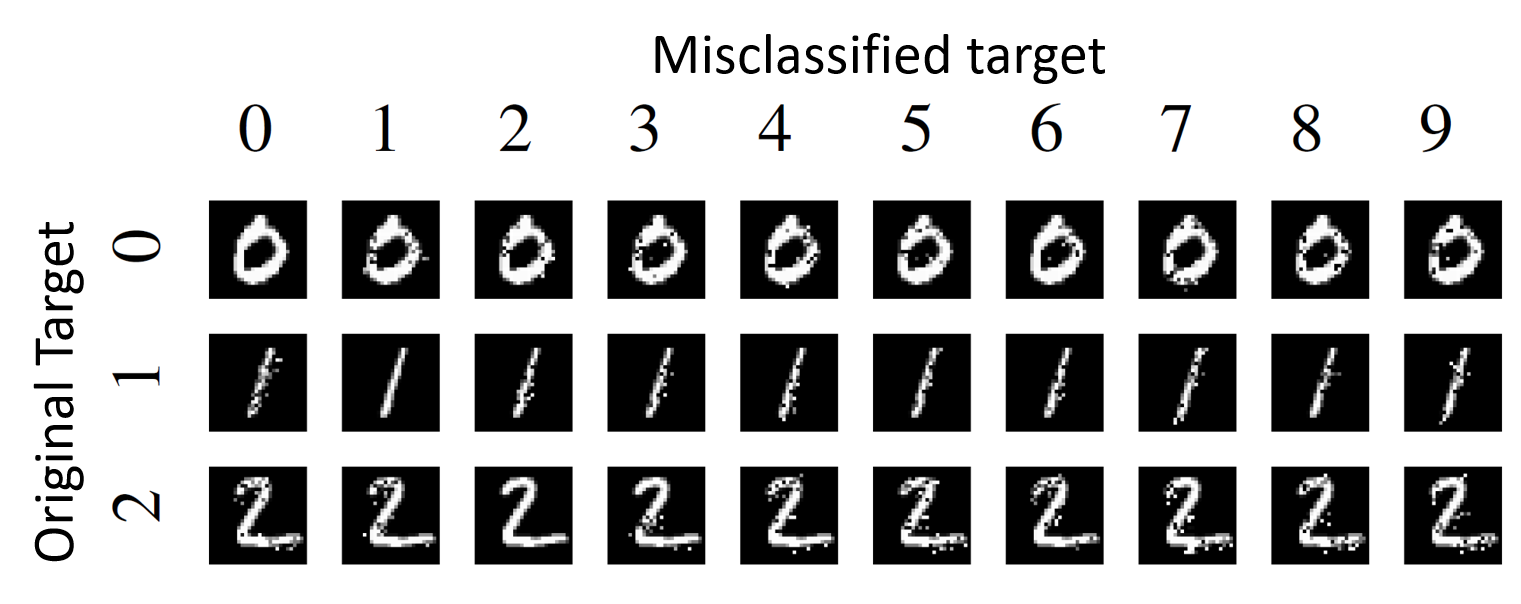}
    \vspace{-1em}
    \caption{Adversarial samples crafted using the $L_0$ attack which are successfully misclassified. The `0', `1', and `2' input samples are taken from the MNIST dataset, and the corresponding adversarial samples is shown under the targeted `wrong' classes. (Samples taken from \cite{carlini_attack})}
    \vspace{-1.5em}
    \label{fig:sample_attack}
\end{figure}

Among all the modern applications of Deep Learning (DL), its impact on Computer Vision (CV) tasks seems ubiquitous. DL networks, specifically the ones based on CNNs, have achieved impressive accuracy in tasks like recognition, segmentation, quality assessment, captioning, enhancement, etc. \cite{krizhevsky2012imagenet, hinton2012deep}. Although DNNs achieve high performance, they come with inherent security risks. Experts in machine learning and security communities have developed sophisticated methods of successfully constructing adversarial inputs using the knowledge of the architecture which can force the model to produce adversary-selected outputs \cite{lbfgs, papernot2016limitations, fgsm}. With simple but imperceptible modifications of a few pixels, a DNN might misclassify the image of handwritten digits \cite{carlini_attack}, as shown in Figure \ref{fig:sample_attack}. Left unattended, such vulnerability could lead to security problems in applications including self-driving cars and internet privacy, potentially causing damages and even threatening human lives.

Unfortunately, no known method has been proposed to entirely eliminate such problems, and few of the proposed methods provide satisfactory improvement. Some of the attempts require substantial changes to the existing network architectures that might limit the capacity of the networks \cite{fgsm, gu2014towards}, and others only provide marginal improvements in robustness against the attacks \cite{bastani2016measuring, huang2015learning, shaham2015understanding}. One method that was proposed recently to be effective against adversarial attacks is defensive distillation \cite{defensiveD}, in which knowledge distillation (KD) is applied on models of the same architecture. \cite{defensiveD} found that KD on models of the same architecture significantly reduces the rate of adversarial sample crafting. 

\begin{figure}[h]
\begin{center}
\includegraphics[width=0.9\linewidth]{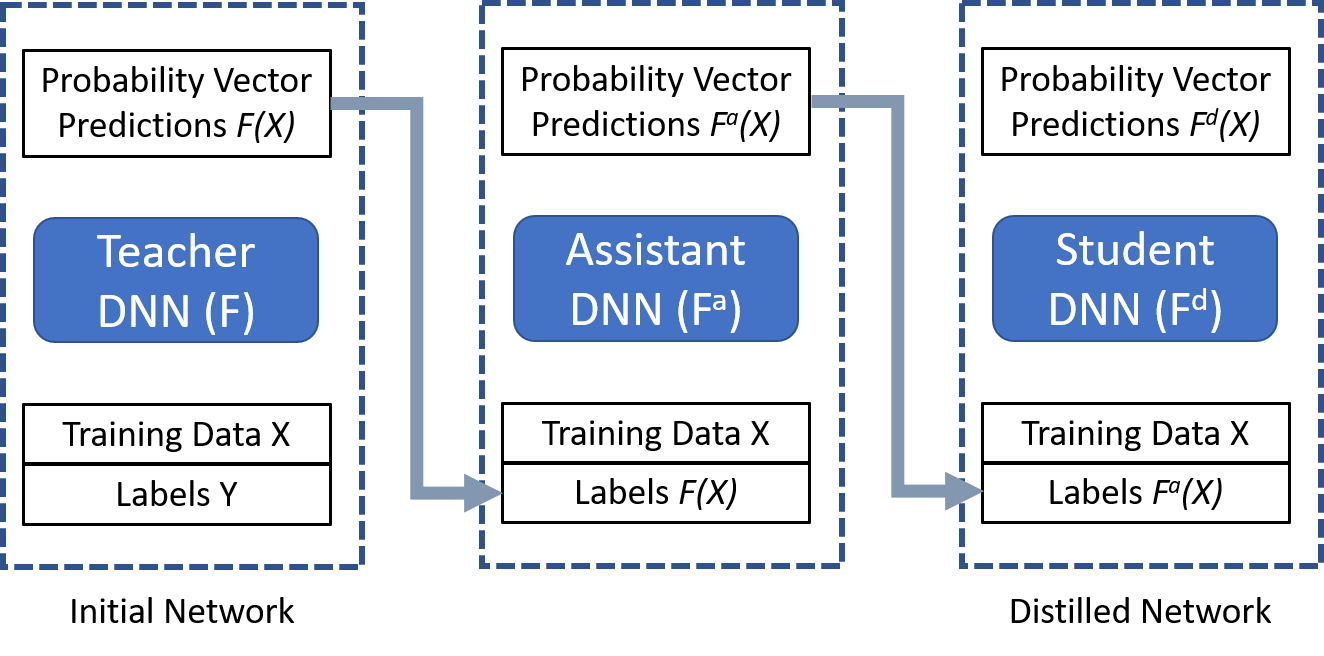}
\vspace{-1em}
\caption{\scriptsize{\textbf{Proposed framework:} In this project, we propose introducing an assistant network in distillation to increase the robustness of student against adversarial perturbations.}}
\vspace{-3em}
\label{fig:framework}
\end{center}
\end{figure}

In this work, we build upon the results from \cite{defensiveD} and explore the usage of multi-step distillation via teacher assistants \cite{TAKD} for defense against adversarial attacks (Fig. \ref{fig:framework}). We also investigate the effect of varying temperature, a hyperparameter that controls the process of the distillation, on such defense. The models have been evaluated in terms of accuracy, sensitivity, and robustness against adversarial attacks.

The rest of this section introduces the two major research our work is built upon: defensive distillation (section \ref{DD}) and teacher assistant knowledge distillation (TAKD, section \ref{TAKD}). Next, we state the method used for crafting adversarial samples (section \ref{sec:attack}). In section \ref{sec:analysis}, \ref{sec:experiments} and \ref{sec:results}, we layout the details of our experiments, the metrics used to evaluate the models, and the obtained results on both MNIST and CIFAR10. Finally, we briefly describe the results of applying distillation with multiple steps in section \ref{sec:multi-step}.

% \todo{add a graph on the first page somewhere? words are boring, maybe a flowchart of DD with TA added}

\subsection{Defensive distillation}\label{DD}
KD was originally introduced as a training procedure that aims to transfer the knowledge learned by a larger network (Teacher) to another relatively smaller one (Student), while maintaining the same level of performance \cite{hinton2015distilling}. The motivation behind this technique is to reduce the computational complexity of some operations, or compression of large networks, such that devices with limited capacity (e.g. smartphones) could benefit from the achievements of deep learning. In KD, the student network is trained not with the original hard binary labels of the dataset, but instead with the soft labels taken from the output probability of the teacher network. 

The technique of defensive distillation introduced by \cite{defensiveD} heavily relies on the intuition behind knowledge distillation (KD) and the common methods of attacks. The intuition behind this procedure is that the knowledge acquired by the teacher network is not only embedded in the weight parameters, but also in the probability vector output by the network. Consider the MNIST digits dataset, where the images of digits 7 and 1 share similar spatial structures. The output of the teacher network thus contains relatively similar probabilities for these two digits. This extra entropy in the probability vector contains structural information about the data compared with the hard binary labels, and thus may be helpful when the student network is faced with adversarial samples that could be ``unnatural", or outside of the data distribution. 

% temperature affects magnitude of probabilities
% One important factor in the distillation procedure is the choice of $T$, the temperature hyperparameter for the softmax operation defined below: 
% \begin{equation}
%     F(X) = \Big[ \frac{e^{z_i(X)/T}}{ \sum_{j=0}^{N-1} e^{z_j(X)/T} } \Big]_{i\in 0 \dots N-1}
% \end{equation}
% where $z_i(X)$ is the output logit of the hidden network for class $i$ in response to input $X$. It is easy to see that the temperature $T$ plays an important role when generating the soft labels from the teacher model with softmax: when $T$ is small, the resulted probability values are more discrete and closer to the hard labels; when $T$ is large, the output is instead more ambiguous, with each class probability close to $1/N$ as $T$ approaches infinity. 

One important factor in the distillation procedure is the choice of distillation temperature ($T$). The effect of the value of $T$ on the resulting defensive distillation is perhaps clearer when we consider the method used for adversarial attacks. Most attacks assume that the adversary has access to the gradient of the network, and can thus estimate the model's sensitivity to the input data (see section \ref{sec:attack}). When the gradients are large, the model is more sensitive because simple perturbations of the input can lead to large changes of the model output, and thus is more prone to attacks. Therefore, a major goal of the defense against adversarial attacks could be to smooth the gradients and to stabilize the model around the input data in the sample space. In this sense, acquiring ``softer" labels from the teacher network (corresponding to larger $T$) should produce more robust student networks, as confirmed by the results from \cite{defensiveD}.

The authors in \cite{defensiveD} found that using defensive distillation reduces the success rate of adversarial sample crafting from 95.89\% to 0.45\% on the MNIST dataset \cite{lecun1998mnist}, and from 87.89\% to 5.11\% on the CIFAR10 dataset \cite{krizhevsky2009learning}.

\subsection{Teacher Assistant Knowledge Distillation (TAKD)} \label{TAKD}
Knowing that defensive distillation could produce student networks that are more robust to adversarial attacks, a natural question one might ask is, what if we apply distillation multiple times? Another motivation is the observation of the teacher-student gap, where the performance of the student is often far than ideal when there is a large difference between the capacity of the student and the teacher.

\cite{TAKD} explored this multi-step distillation strategy in normal KD settings with the task of image recognition. Instead of training one student directly from the teacher, this method employs a teacher assistant, a network that has an architecture of intermediate size between the teacher and the student, and serves as an intermediate step in the distillation process. In other words, the TA is first trained with the soft labels from the teacher, and then produces another set of soft labels to train the student. The authors observed that student models trained with TAs outperform those trained directly with the teacher. They also discuss the choice of TA size and the distillation path for multi-step distillation beyond one TA. They observed that the more TAs employed in between distillations, the better the student.

In this work, we combine TAKD with defensive distillation, and explore whether adding intermediate steps in the distillation process can lead to even larger improvement in the robustness of the student model.

\section{Adversarial Attacks}\label{sec:attack}
The main goal of an adversarial attack algorithm is to produce a perturbation small enough to be imperceptible to human eyes, but large enough to produce misclassification. For a sample $X$ and a trained classifier $F$, and adversarial attack produces an input sample $X^* = X+ \delta X$ (where $\delta X$) is the perturbation, such that $F(X^*) = Y^*$ and $Y^* \neq Y$. There have been several attack algorithms based on the $L_0$, $L_2$, and $L_\infty$ distance metrics proposed in literature such as box-constrained L-BFGS method \cite{lbfgs} and Deepfool \cite{deepfool} based on $L_2$, Fast Gradient Sign \cite{fgsm} and Iterative Gradient Sign \cite{igsm} methods based on $L_\infty$ distance, and Jacobian-based Saliency Map Attack (JSMA) \cite{papernot2016limitations} based on the $L_0$ distance. But the methods we used in our project were proposed in \cite{carlini_attack} which improve upon the mentioned works. The adversarial attack optimization problem is formally defined as:

\begin{equation}\begin{array}{ll}
\operatorname{minimize} & \mathcal{D}(X, X+\delta X) \\
\text { such that } & F(X+\delta X)=t \\
& X+\delta X \in[0,1]^{n}
\end{array}\end{equation}

Here $F$ is the classification model, $t$ is the targeted `wrong' class, and $\mathcal{D}$ is the distance metric. The constraint $F(X+\delta X)$ is highly non-linear, so the problem is expressed differently for applying optimization algorithms. An objective function $f$ is defined such that $F(X+\delta X) = t$ iff $f(X+\delta X) \leq 0$. The authors have proposed and analyzed several such $f$ in their paper. Replacing the first constraint, an alternative formulation of the optimization problem is given as:

\begin{equation}\begin{array}{l}
\text { minimize } \mathcal{D}(X, X+\delta X)+c \cdot f(X+\delta X) \\
\text { such that } X+\delta X \in[0,1]^{n}
\end{array}\end{equation}

Here $\mathcal{D}(X, X+\delta X) = \lVert \delta X \rVert_p$, if we use the $L_p$ norm. 

\subsection{$L_2$ Attack}
Instead of optimizing for $\delta X$, the objective is optimized over a transformed variable $w$, s.t. 
\begin{equation}
\delta X_{i}=\frac{1}{2}\left(\tanh \left(w_{i}\right)+1\right)-X_{i}
\end{equation}
As, $-1 \leq \tanh (w_i) \leq 1$ implies $0 \leq X_i + \delta X_i \leq 1$. If $X$ is the input sample, and $t$ is the chosen `wrong' target class ($ t \neq F(X)$), then the attack finds $w$ solving

\begin{equation}
\text { minimize }\left\|\frac{1}{2}(\tanh (w)+1)-X\right\|_{2}^{2}+c \cdot f\left(\frac{1}{2}(\tanh (w)+1)\right) \\
\end{equation}
\begin{equation}
f\left(X^{*}\right)=\max \left(\max \left\{Z\left(X^{*}\right)_{i}: i \neq t\right\}-Z\left(X^{*}\right)_{t},-\kappa\right)
\end{equation}
Here $f$ is the best suited objective function proposed in the paper, and $\kappa$ controls the confidence of misclassification. It encourages the solver to find an adversarial sample $X^*$ which will be classified as class $t$ with high confidence. For the purpose of experiments in this project, $\kappa$ has been set to zero. To avoid getting stuck at a local minimum, gradient descent is run with multiple random starting points close to $X$ for a fixed number of iterations. The random points are sampled at random from a norm ball of radius $r$ centered at $X$, where $r$ is the closest adversarial sample found so far.

\subsection{$L_0$ Attack}
The $L_0$ metric is non-differentiable and therefore standard gradient descent cannot be applied. The $L_0$ attack algorithm uses the $L_2$ adversary to eliminate pixels that are unimportant. Starting with the original image $X$, let $\delta X$ be the solution found by $L_2$ adversary, such that $X^ = X + \delta X$ is an adversarial sample. The gradient of the objective function is computed ($g = \nabla f(X*)$). The pixel with the minimum gradient ($i = \operatorname{arg\min}_i g_i \cdot \delta X_i$) is removed from the allowed set, as it will have the least impact on the output. On each iteration, pixels are eliminated and the adversary is restricted to modify only the allowed set. This process is repeated until the $L_2$ adversary fails. By process of elimination, a final subset of important pixels can be identified. The constant $c$ is initially set to a very low value, and $L_2$ attack is run using this value. Upon failure, the value is doubled and the attack is run again till $c$ exceeds a threshold, which is declared as a failure. At each iteration, the solution found in the previous one is used as the starting point for gradient descent, and as such, the attack algorithm is much more efficient than previous $L_0$ attacks in the literature.

\subsection{$L_\infty$ Attack}
The $L_\infty$ distance is not fully differentiable and the authors observed that the performance was poor when using the basic objective function. They also noticed that as the $\left\|\delta X \right\|_\infty$ penalizes the largest element, gradient descent gets stuck oscillating between two suboptimal options. To circumvent this, the $L_\infty$ term is replaced in the objective function with apenalty for any element of $\delta X$ that exceeds threshold $\tau$. This prevents the oscillation in gradient descent, as all the large values are penalized. The resulting objective function is 
\begin{equation}
\text { minimize } \quad c \cdot f(X+\delta X)+\cdot \sum_{i}\left[\left(\delta X_{i}-\tau\right)^{+}\right]
\end{equation}
The threshold $\tau$ is initially set to $1$, and decreased in each iteration. If after an iteration, all of the components of $\delta X$ are less than $\tau$, it is reduced by a factor of $0.9$ and repeated. The constant $c$ is chosen similarly to the $L_0$ case. Initially set to a low value, and if the $L_\infty$ adversary fails, the value of $c$ is doubled until it exceeds a threshold, when it is declared as a failure. Similar to $L_0$ adversary, gradient descent at each iteration is performed with a ``warm-start" for efficiency. 

\section{Analysis of Defensive Distillation}\label{sec:analysis}
In defensive distillation, two networks of similar architecture, the teacher (or primary) and the student (or distilled), are trained sequentially. The input to the distillation process is a set of sample images $X \in \mathcal{X}$ and the corresponding `hard' labels $Y(X)$, which is a set of one-hot vectors corresponding to the correct class. Given the training set $\{(X,Y(X)) \ X \in \mathcal{X}\}$, the primary teacher $F$ is trained with a softmax layer of temperature $T$. $F(X)$ is the resulting probability vector computed for input $X$ using the trained teacher network, and these are used as the `soft' labels for training the distilled student network. If $F$ has parameters $\theta_F$, then $F(X) = p(.|X,\theta_F)$ is the probability distribution of the output. The new training set $\{(X,F(X)) \ X \in \mathcal{X}\}$ is used to train the student DNN ($F^d$) which has the same neural architecture and the same softmax temperature $T$ as its teacher. Using soft-targets for training $F^d$ imparts additional knowledge found in probability vectors than compared to hard labels. The additional entropy encodes the relative differences between classes. This relative probabilistic information prevents the network from fitting too tightly to the data and increases its generalizability. Inspired by TAKD, we propose and show that adding an assistant model $F^a$ of the same architecture and softmax temperature in between the teacher and distilled student model increases robustness to adversarial attacks. 
\subsection{Impact on Training}
For training the DNN classifier $F$, the training algorithm minimizes the empirical risk:
\begin{equation}
\arg \min _{\theta_{F}}-\frac{1}{|\mathcal{X}|} \sum_{X \in \mathcal{X}} \sum_{i \in 0 . . N} Y_{i}(X) \log F_{i}(X)
\end{equation}
So, to decrease the log-likelihood $\ell(F, X, Y(X))=-Y(X) \cdot \log F(X)$ of $F$ on $(X,Y(X))$, the optimizer adjusts $\theta_F$ so that $F(X)$ can get close to $Y(X)$. As, $Y(X)$ is a one-hot vector, the optimization problem can be written as 
\begin{equation}
\arg \min _{\theta_{F}}-\frac{1}{|\mathcal{X}|} \sum_{X \in \mathcal{X}} \log F_{t(X)}(X)
\end{equation}
So, the optimizer effectively pushes the model to make overconfident predictions on $t(X)$, while pushing other probabilities to zero. In contrast, when the student network is trained using soft labels $F(X)$, the optimization problem is given by:
\begin{equation}
\arg \min _{\theta_{F}}-\frac{1}{|\mathcal{X}|} \sum_{X \in \mathcal{X}} \sum_{i \in 0 . . N} F_{i}(X) \log F_{i}^{d}(X)
\end{equation}
Using probabilities $F_j(X)$ ensures that the training algorithm constrains the output neurons $F_j^d(X)$ proportionally to their likelihood when updating $\theta_F$. This ensures that the model learns the relative likelihood among classes and is not forced into hard predictions, this increasing its generalizability. Introducing $F^a$ in between them further increases this trend. Ideally, with enough samples and training, $F^a$ would eventually converge to $F$, and consequently $F^d$ would converge to $F^a$, but empirically the model robustness is seen to increase with distillation.

\subsection{Model Sensitivity}
The training procedure gives an intuition about the higher generalizability of the distilled models, but a sensitivity metric shows how crafting adversarial samples gets harder with increasing the distillation temperature. As discussed before, adversarial attacks identify inputs that have a higher chance of misclassification with lower perturbations. In other words, models that are more sensitive to perturbations will cause a higher change in output with small input variations. The authors in \cite{defensiveD} have shown that the sensitivity of a model with respect to its input variations is dictated by the magnitude of its Jacobian or gradients. The expression for the $(i,j)$ component of the Jacobian for model $F$ at distillation temperature $T$ is:
\begin{align}
\left.\frac{\partial F_{i}(X)}{\partial X_{j}}\right|_{T} & =\frac{\partial}{\partial X_{j}}\left(\frac{e^{z_{i}(X) / T}}{\sum_{l=0}^{N-1} e^{z_{l}(X) / T}}\right) \\
& =\frac{1}{T} \frac{e^{z_{i} / T}}{g^{2}(X)}\left(\sum_{l=0}^{N-1}\left(\frac{\partial z_{i}}{\partial X_{j}}-\frac{\partial z_{l}}{\partial X_{j}}\right) e^{z_{l} / T}\right)
\end{align}
Here $z_0(X),...,Z_{N-1}(X)$ are the logits. For fixed values of logits, increasing distillation temperature $T$ will lead to decreasing the magnitude of the model gradients as can be inferred from the above expression. This reduces the model sensitivity to adversarial perturbations, and larger input variations are needed to craft the samples. In other words, increasing distillation temperature reduces the magnitude of the adversarial gradients. It is to be noted that the distillation temperature is introduced only during training, and it is not used in the evaluation, i.e. for testing the models, $T$ is set to 1.  

\subsection{Model Robustness}
To measure the efficacy of defensive distillation, the robustness metric was introduced in \cite{Fawzi2017AnalysisOC}. The robustness of a model refers to its resistance to adversarial perturbations. A robust DNN should perform well (in terms of prediction accuracy) both on the training and the outside data, and should be consistent in its class predictions for inputs in the neighborhood of a given sample. Robustness is achieved when this consistency can be ensured in a closed neighborhood. The larger this neighborhood, the higher the robustness of the DNN. The neighborhood cannot be extended indefinitely, as that is the case of a constant function. The robustness of a trained DNN classifier $F$ is given as:
\begin{equation}
\rho_{a d v}(F)=E_{\mu}\left[\Delta_{a d v}(X, F)\right]
\end{equation}
$X$ is the input data drawn from the true distribution $\mu$, and $\Delta_{a d v}(X, F)$ is the minimum perturbation that causes a misclassification. Therefore,
\begin{equation}
\Delta_{a d v}(X, F)=\arg \min _{\delta X}\{\|\delta X\|: F(X+\delta X) \neq F(X)\}
\end{equation}
The distance metric can be chosen accordingly. For our project, we compute the average value of $\Delta_{a d v}(X, F)$ computed over all test samples. Higher the value, more robust is the model.

\section{Experiments}\label{sec:experiments}

\subsection{Dataset}
For our project, we used two legacy image classification datasets - MNIST \cite{lecun1998mnist} and CIFAR10 \cite{cifar}. The MNIST dataset is used for classifying handwritten digits (0-9). It consists of 60,000 training samples and 10,000 testing samples, and the pixels are encoded to $[0,1]$. The CIFAR10 dataset consists of 60,000 color images (three color components) divided into 50,000 training and 10,000 testing samples. Similar to MNIST, CIFAR10 images are classified into 10 mutually exclusive classes. We chose these simple datasets because we could construct relatively shallow models to achieve satisfactory performance. Also, we needed to train and evaluate a large number of models for different temperatures and multiple steps, so we wanted to work with small datasets that can be trained efficiently. Moreover, the previous papers on defensive distillation also dealt with these two simple multi-class classification problems, and we wanted to demonstrate improvements over them.

\subsection{Model Architecture and Training}
To remain consistent with previous works on defensive distillation, we used the same models as \cite{defensiveD} and \cite{carlini_attack}. Both the DNN architectures have 9 layers consisting of 4 convolution layers, 2 max pooling layers, and 3 fully connected dense layers in the head. Momentum and parameter decay were used to ensure convergence, and dropout was used to prevent overfitting. For the softmax layer, we experimented on distillation temperatures $T = \{1,2,5,10,20,30,40\}$. For training, we used a batch size of 128, and a learning rate of $\eta = 0.01$, decay rate $10^{-6}$, momentum 0.9, and SGD optimizer for 50 epochs. 

\subsection{Attack Parameters}
We experimented with the three attacks proposed in \cite{carlini_attack} as described in section \ref{sec:attack}. For each of the attacks, we evaluated the robustness (average perturbation) of the teacher, assistant, and student models for the entire temperature range. For $L_2$ attack, we used a learning rate of $10^{-2}$, maximum number of iterations as 10,000, initial value of constant $c$ as $10^{-3}$, and zero confidence value. For $L_0$ attack, we reduced the maximum number of iterations to 1,000, and the upper limit of the constant to $2\times10^{-6}$. For $L_\infty$ attack, the learning rate was fixed at $5\times10^{-3}$, the initial value of the constant was set to $10^-5$, and the upper threshold was set to $20$. The number of maximum iterations was kept the same as $L_0$ attack.

\section{Results}\label{sec:results}
\subsection{Accuracy}
\begin{figure}[h]
\vspace{-1em}
\begin{center}
\includegraphics[width=1\linewidth]{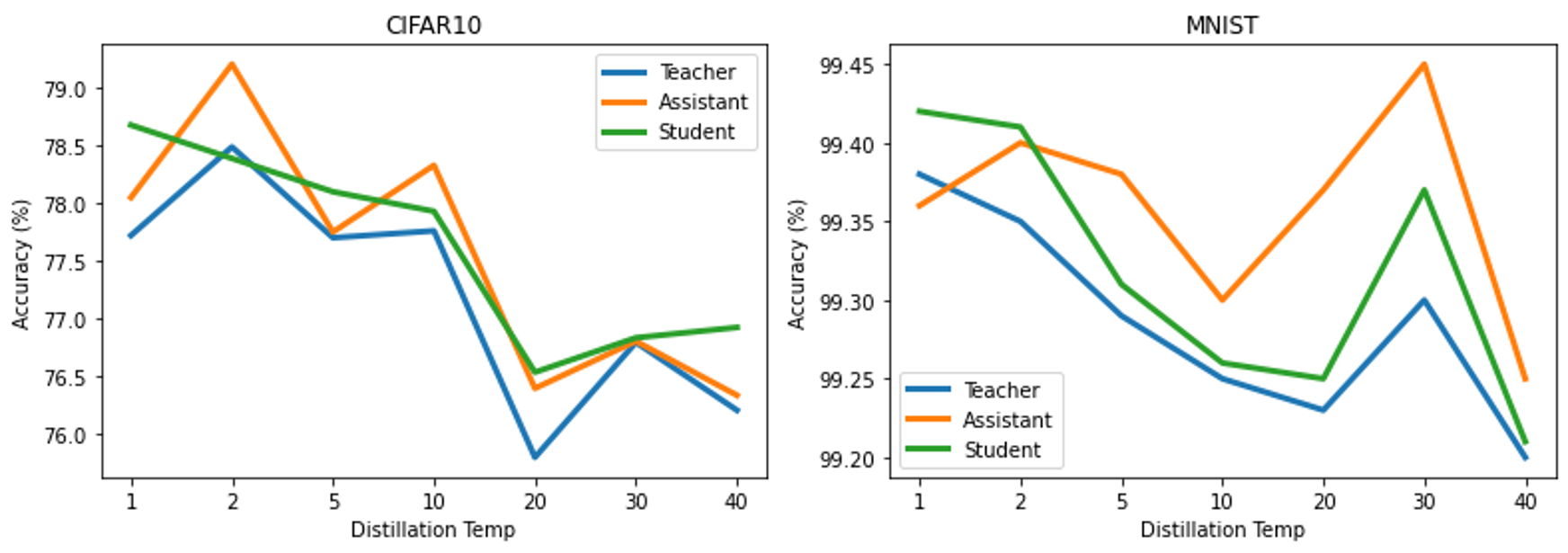}
\vspace{-2em}
\caption{\scriptsize{\textbf{Influence of distillation on accuracy:} The accuracy of the teacher, assistant, and student models for different distillation temperatures evaluated on the test datasets.}}
\vspace{-2em}
\label{fig:model_acc}
\end{center}
\end{figure}

In the first set of experiments, we wanted to observe the variation of the accuracy of the distilled models as compared to the original teacher model when trained at different distillation temperatures. For each of the datasets, MNIST and CIFAR10, we evaluate the performance of the models on the test dataset (10,000 images). As mentioned before, for testing we keep the distillation temperature to be $1$. The baseline accuracy for both the datasets are measured by evaluating the teacher model trained with temperature $T=1$. The accuracy for baseline MNIST model $F_{MNIST}$ was $99.38\%$, and for $F_{CIFAR10}$ was $77.72\%$. From figure \ref{fig:model_acc} we can notice that the accuracy for either of the datasets varies very little with distillation. For MNIST, the maximum variation that can be noticed w.r.t the teacher is about $0.15\%$, whereas for CIFAR10, the maximum variation noticed is $1.2\%$. It should also be noted that for each of the models, there is a trend of accuracy decreasing with temperature. With the increase in temperature, most of the class outputs have higher probability values and that might make the class distinctions a bit difficult (As $T \to \infty$, the output of softmax converges to $1/N$). It is also interesting to note that in most cases, for a fixed temperature, the distilled models leads to better performance. This may be due to the increase in generalization ability of the network that leads to better prediction on unobserved samples.

\subsection{Sensitivity}
\begin{figure}[h]
\vspace{-1.5em}
\begin{center}
\includegraphics[width=0.8\linewidth]{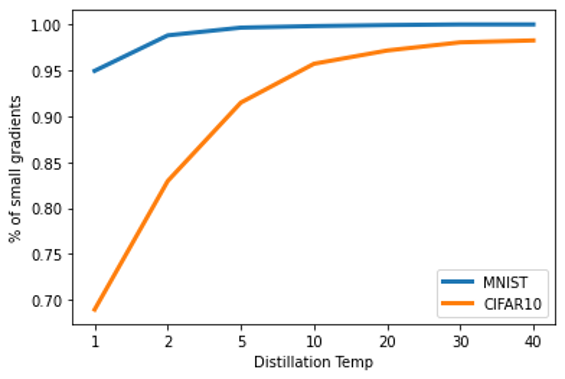}
\vspace{-1.5em}
\caption{\scriptsize{\textbf{Influence of distillation temperature on model sensitivity:} The mean value of the gradient amplitudes calculated for 10,000 test samples for different distillation temperatures are computed, and the proportion of small gradients are calculated. As we expected, higher temperature causes more gradients to go to very small values, effectively smoothing the network output.}}
\vspace{-2em}
\label{fig:model_grads}
\end{center}
\end{figure}

The second set of experiments pertains to the effect of distillation on the sensitivity of the models. As described in Section \ref{sec:analysis}, the sensitivity of a model is measured using the magnitude of its gradients. For each of the two datasets, we computed the gradient magnitudes for the 10,000 test samples as inputs, and took their average. According to the hypothesis presented in the analysis, increasing the distillation temperature should result in decreasing the magnitude of the gradients. Small gradients mean the function is smoother around the sample points in the distribution. A smoother function with lower gradient magnitudes would need higher perturbation for misclassification, which we will show later. The effect of temperature is illustrated in figure \ref{fig:model_grads}. For the teacher models, each trained at different temperatures, we compute the magnitude of the gradients averaged over all input dimensions. Repeating this for 10,000 test samples, we have 10,000 instances of mean gradient amplitudes. We then calculate the proportion of samples for which the gradient magnitude is close to zero. For simplifying the illustration, we consider any gradient with amplitude $<10^{-10}$ to be close to zero. The proportion of such gradients is plotted for both models. As expected, higher temperatures cause the trained models to produce smaller gradients and thus smoother outputs.

\subsection{Robustness}
Robustness is measured by computing the minimum perturbation needed for misclassification (Sec \ref{sec:analysis}). In the third set of experiments, we evaluate the efficacy of robustness against the attacks stated in section \ref{sec:attack}. Our goal is to observe how temperature affects the robustness of a model, and if the introduction of the assistant in distillation improves the robustness of the student model. To evaluate this, we craft adversarial samples using each attack on all 10,000 inputs from the test set. For each sample input, the minimum perturbation for misclassifying it is measured (distance between the input and the crafted sample). Finally, we calculate the average of the perturbations over the entire test set which is the robustness value of the model. We repeat this experiment for both the MNIST and CIFAR10 datasets. 

\begin{figure}[h]
\vspace{-1em}
\begin{center}
\includegraphics[width=1\linewidth]{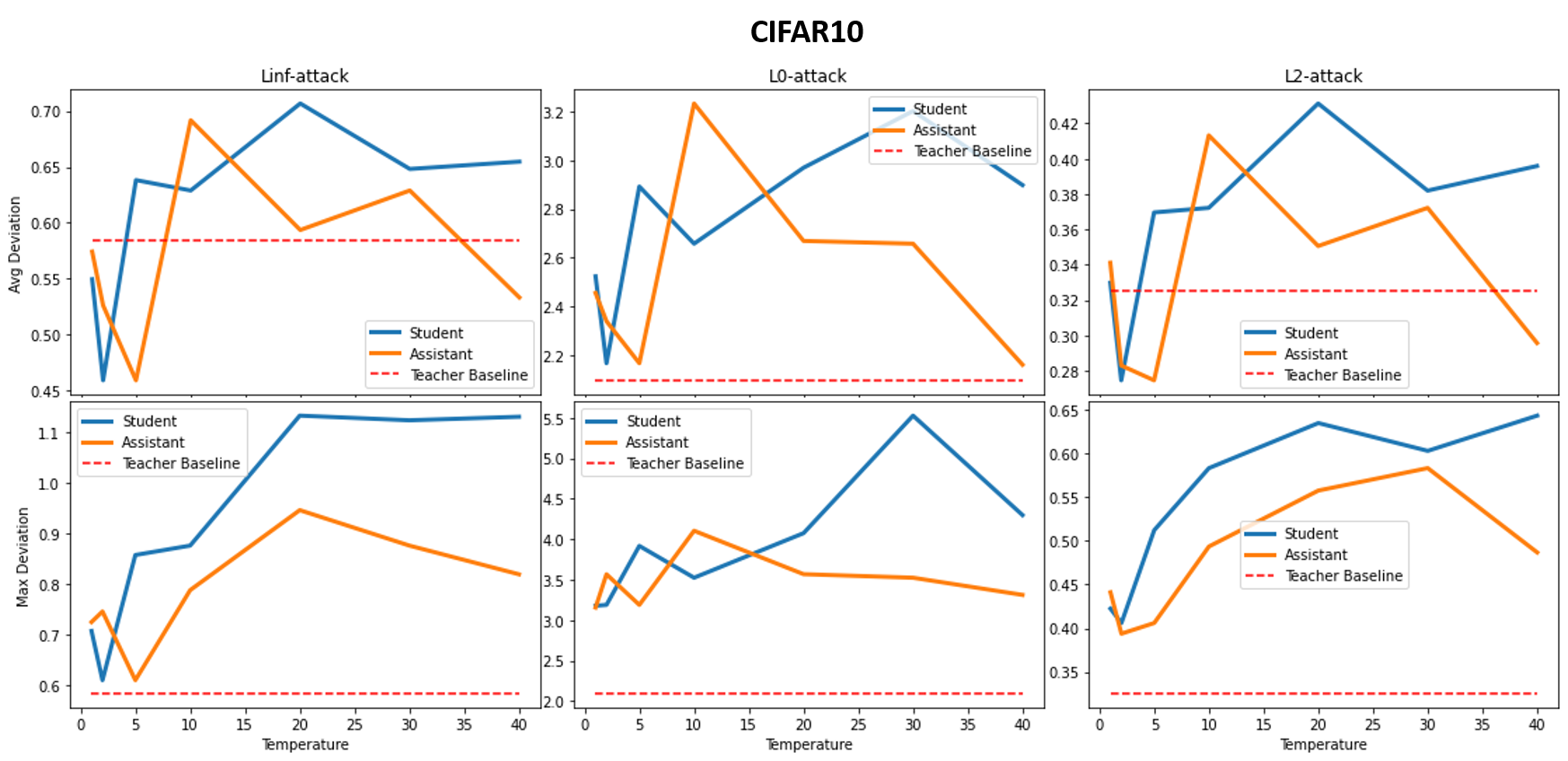}
\vspace{1em}
\includegraphics[width=1\linewidth]{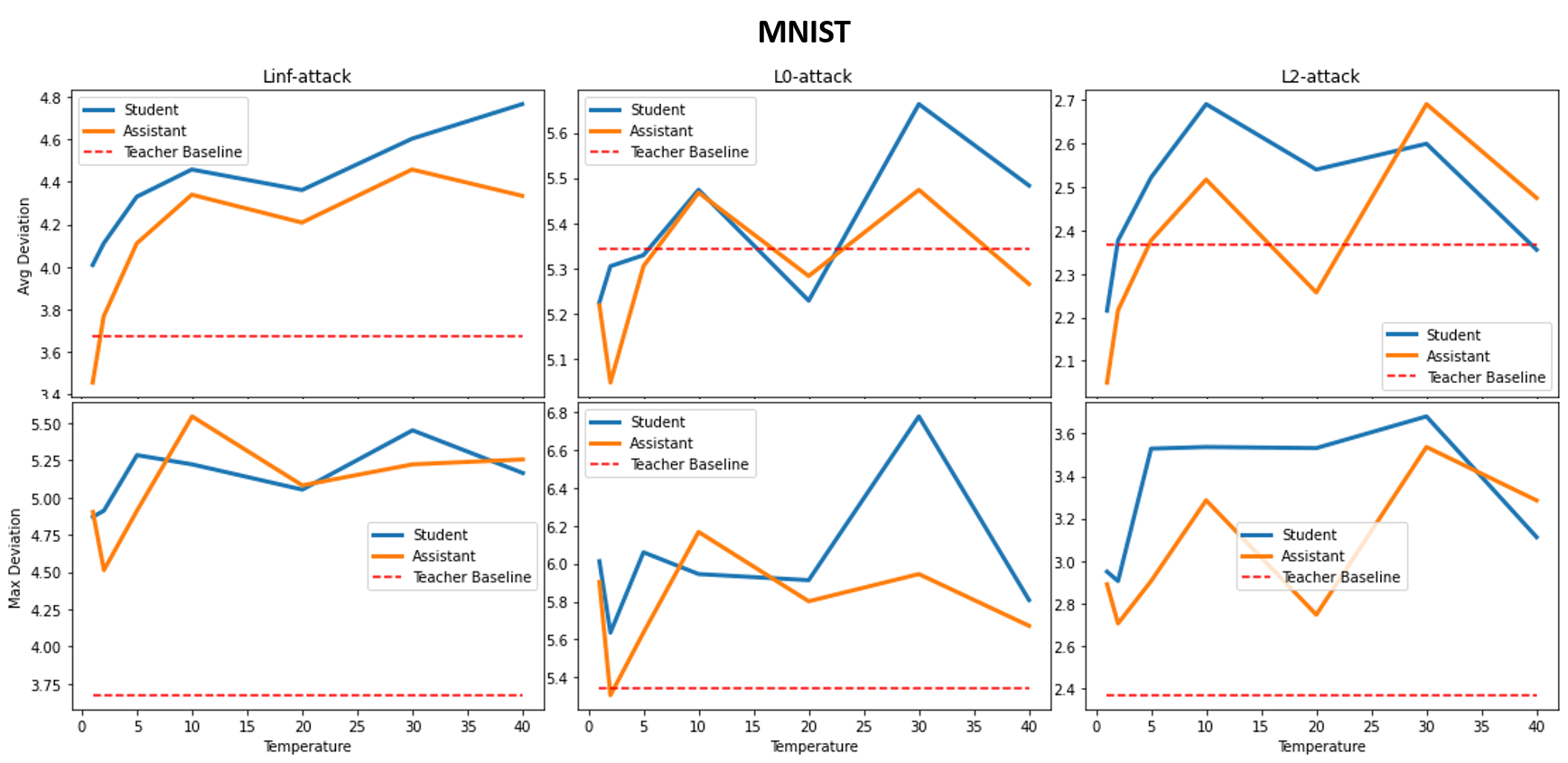}
\vspace{-2.5em}
\caption{\scriptsize{\textbf{Influence of temperature and distillation on robustness:} The plots show the magnitude of average perturbation or robustness (top) and max perturbation (bottom) for each of the two datasets. It can be observed that generally robustness increases with distillation temperature and the introduction of the assistant (orange) increases the robustness of the student (blue).}}
\label{fig:model_robust}
\vspace{-1.5em}
\end{center}
\end{figure}

Figure \ref{fig:model_robust} is a collection of plots for different models. In each subplot, the orange line indicates the variation of adversarial perturbation for the assistant model, the blue line indicates that of the student model, and the red dotted line is that of the baseline teacher model. For each dataset, the plots on the top row are of the mean deviations (or perturbations), which is the robustness metric, and the bottom row is the maximum deviation required to craft an adversarial sample. Irrespective of the attack, it can be observed that, in general, increasing the distillation temperature improves the robustness of the model. Also, on average, the student model has higher robustness than the teacher and assistant models. This shows that introducing an assistant network did improve the robustness against adversarial attacks in most cases. Among the different attacks, it can be noticed that $L_2$ is the most efficient as it generates adversarial samples successfully with the least perturbations among the three, whereas $L_0$ is the least efficient. From these sets of experiments, it can be inferred that an assistant model in distillation does help in defending better against adversarial attacks with effectively very minimal hit to performance. 

\subsection{Confidence}
\begin{figure}[h]
\begin{center}
\includegraphics[width=0.8\linewidth]{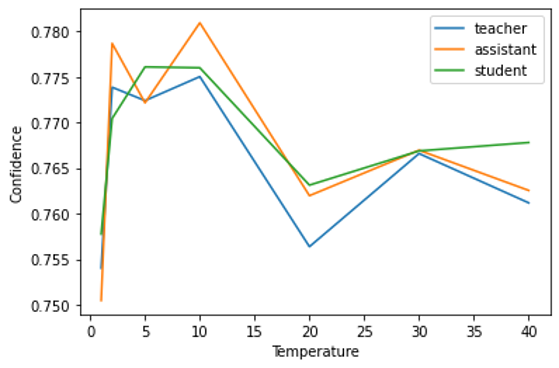}
\vspace{-1em}
\caption{\scriptsize{\textbf{Influence of distillation on confidence:} The confidence values of the CIFAR10 models computed over the test set.}}
\vspace{-2em}
\label{fig:cifar_conf}
\end{center}
\end{figure}

The final experiments that we conducted were computing the confidence values of the previously stated models. The confidence of a model calculated over a dataset $\mathcal{X}$ is the average of the following quantity over all $X \in \mathcal{X}$:
\begin{equation}
C(X)=\left\{\begin{array}{c}
0 \text { if } \arg \max _{i} F_{i}(X) \neq t(X) \\
\arg \max _{i} F_{i}(X) \text { otherwise }
\end{array}\right.
\end{equation}
Here $t(X)$ is the correct class for input $X$. In \cite{defensiveD}, the authors have stated the confidence values increase with distillation temperature for the models trained on CIFAR10. We repeated the same experiment for our models but did not observe any concrete trend over temperature. We also did not observe any substantial difference among the teacher, assistant, and student models as shown in Fig. \ref{fig:cifar_conf}. The variation in confidence values is very little in either case.

\section{Multi-step Distillation}\label{sec:multi-step}
\begin{figure}[h]
\vspace{-1em}
\begin{center}
\includegraphics[width=1\linewidth]{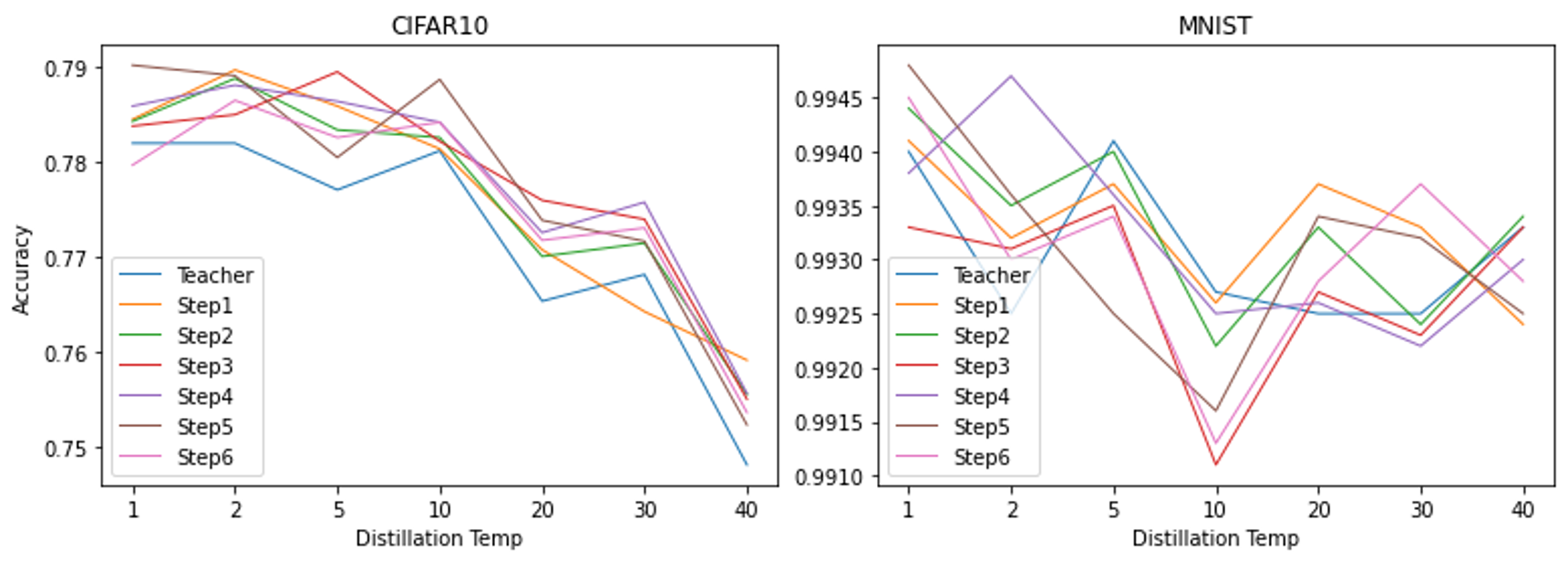}
\vspace{-2em}
\caption{\scriptsize{\textbf{Model accuracies for multi-step distillation:} The performance of the MNIST and CIFAR10 models for different levels of distillation and increasing temperature.}}
\vspace{-1em}
\label{fig:acc_multistep}
\end{center}
\end{figure}

As we noticed an improvement in robustness after introducing an assistant in defensive distillation, we wanted to observe the effect of multi-step distillation. Specifically, for the experiments in this section, six models were trained sequentially after the teacher, and the output of each was used as the input labels for the next one. First, we study the effect of multi-step distillation on the classification performance of the models. Fig. \ref{fig:acc_multistep} plots the model accuracies for different distillation steps and varying the temperature. We can observe that the variation in performance is very little, although we can see the temperature trend in the case of CIFAR10 models as seen before. Also, increasing the levels of distillation does not have a significant effect on accuracy for either of the model classes.

\begin{figure}[h]
\begin{center}
\includegraphics[width=1\linewidth]{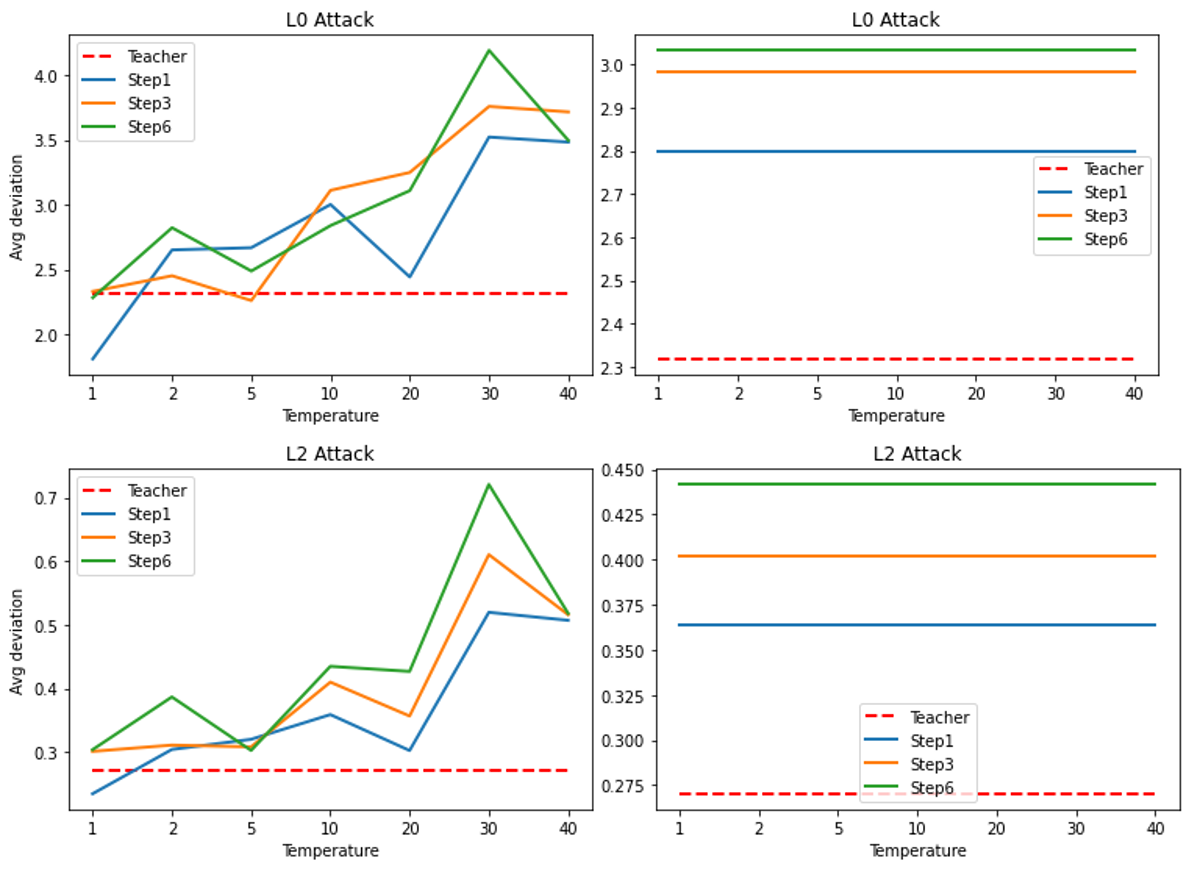}
\vspace{-1.5em}
\caption{\scriptsize{\textbf{Robustness for multi-step distillation:} The robustness of the CIFAR10 models for different levels of distillation and increasing temperature.}}
\vspace{-2em}
\label{fig:robust_multistep}
\end{center}
\end{figure}

Testing the robustness of multi-step distilled models offers some interesting insights. Fig. \ref{fig:robust_multistep} is plotted using the robustness metric calculated for different distilled models at varying temperatures. For the sake of clarity, we only show the results of three steps of distillation. The figures on the left show the increasing trend of robustness with increasing temperature as shown before. The figures on the right show the robustness value averaged over all temperatures. These figures clearly show that robustness against the attacks increases as we carry out distillation through multiple steps. Although the improvement is not huge, but it is still noticeable. We only show the plots for $L_0$ and $L_2$ attacks on CIFAR10 models, but the same can be shown for other models and attacks as well. In conclusion, multi-step distillation does help in making the model more robust against attacks with an insignificant hit to performance. 

\section{Conclusion}
In this project we build on the work in defensive distillation \cite{defensiveD}, and inspired by the success of TAKD \cite{TAKD}, we proposed that introducing the assistant model in distillation would improve the robustness of the student against adversarial attacks. We use the state-of-the-art adversarial attacks proposed in \cite{carlini_attack} and test it on our models trained on CIFAR10 and MNIST datasets. For both the datasets, we verify the claims on robustness and sensitivity of the distilled models, and also provide empirical evidence supporting our proposed hypothesis. We further experiment on multi-step distillations, and successfully show that with the increase of distillation levels, models tend to get more robust with very little hit to performance. However, we acknowledge that the field of defensive distillation is in its infancy and it is very difficult to analytically prove these claims. Also, modern adversarial attacks are very effective and for most cases distillation cannot provide substantial defense. In the future, we would like to delve deeper into the analysis and try to mathematically support our statements. We would also like to experiment with more architectures and attacks to generalize our claims. Defensive distillation does show promise but still needs to be studied extensively before implementing in real-world applications.

% \begin{table}[htbp]
% \caption{Table Type Styles}
% \begin{center}
% \begin{tabular}{|c|c|c|c|}
% \hline
% \textbf{Table}&\multicolumn{3}{|c|}{\textbf{Table Column Head}} \\
% \cline{2-4} 
% \textbf{Head} & \textbf{\textit{Table column subhead}}& \textbf{\textit{Subhead}}& \textbf{\textit{Subhead}} \\
% \hline
% copy& More table copy$^{\mathrm{a}}$& &  \\
% \hline
% \multicolumn{4}{l}{$^{\mathrm{a}}$Sample of a Table footnote.}
% \end{tabular}
% \label{tab1}
% \end{center}
% \end{table}

% \begin{figure}[htbp]
% \centerline{\includegraphics{fig1.png}}
% \caption{Example of a figure caption.}
% \label{fig}
% \end{figure}

% \cite{TAKD} \cite{defensiveD} \cite{DBLP:journals/corr/abs-1902-03393}

\bibliographystyle{ieeetr}  
\bibliography{ref}

\end{document}